\documentclass[10pt,twocolumn,letterpaper]{article}

\usepackage{cvpr}
\usepackage[]{authblk}
\usepackage{times}
\usepackage{epsfig}
\usepackage{graphicx}
\usepackage{amsmath}
\usepackage{amssymb}
\usepackage{bm}
\usepackage{multirow}
\usepackage[tight,footnotesize]{subfigure}
\usepackage[marginal]{footmisc}
\usepackage{balance}\usepackage{graphicx}
\usepackage{latexsym, amsfonts, amsmath, amsthm, amssymb, mathrsfs, bm}
\usepackage{multirow}
\usepackage{algorithm}
\usepackage{algorithmic}
\usepackage{multirow}
\usepackage[tight,footnotesize]{subfigure}

\usepackage{threeparttable}
\usepackage{amsmath}
\usepackage{dcolumn}
\usepackage{multirow}
\usepackage{booktabs}

\usepackage[breaklinks=true,bookmarks=false]{hyperref}

\cvprfinalcopy % *** Uncomment this line for the final submission

\def\cvprPaperID{****} % *** Enter the CVPR Paper ID here

\begin{document}

%%%%%%%%% TITLE
\title{Person Re-Identification Meets Image Search}

\author[1]{Liang Zheng*}
\author[1]{Liyue Shen*}
\author[1]{Lu Tian*}
\author[1]{Shengjin Wang}
\author[1]{Jiahao Bu}
\author[2]{Qi Tian}
\setlength{\affilsep}{0em} \affil[1]{Tsinghua University, Beijing 100084, China}
\affil[2]{University of Texas at San Antonio, TX 78249, USA
\authorcr{\tt\small liangzheng06@gmail.com \tt\small qitian@cs.utsa.edu}}

\maketitle
%\thispagestyle{empty}

%%%%%%%%% ABSTRACT
\begin{abstract}
For long time\footnote{\noindent{* Three authors contribute equally to this work.}}, person re-identification and image search are two separately studied tasks. However, for person re-identification, the effectiveness of local features and the ``query-search'' mode make it well posed for image search techniques.

%State-of-the-art person re-identification systems typically rely on local feature matching with prior geometric constraints. However, the perfect alignment of hand-drawn bounding boxes are not available in realistic settings, and in this scenario detector misalignment would destroy what we have achieved so far.

In the light of recent advances in image search, this paper proposes to treat person re-identification as an image search problem. Specifically, this paper claims two major contributions. 1) By designing an unsupervised Bag-of-Words representation, we are devoted to bridging the gap between the two tasks by integrating techniques from image search in person re-identification. We show that our system sets up an effective yet efficient baseline that is amenable to further supervised/unsupervised improvements. 2) We contribute a new high quality dataset which uses DPM detector and includes a number of distractor images. Our dataset reaches closer to realistic settings, and new perspectives are provided.

Compared with approaches that rely on feature-feature match, our method is faster by over two orders of magnitude. Moreover, on three datasets, we report competitive results compared with the state-of-the-art methods.
\end{abstract}

%%%%%%%%% BODY TEXT
\section{Introduction}
This paper considers the task of person re-identification. Given a probe image (query), our task is to search in a gallery (database) for images that contain the same person. Person Re-identification has important applications in video surveillance, \eg, cross-camera visual tracking, multi-camera event detection, \emph{etc}.
%The core component of person re-id consists in matching pedestrians captured in different camera views with visual features.
This task still remains an unsolved problem, due to the difficulty in visual matching caused by the extensive variations in illumination, viewpoint, pose, photometric settings of cameras, low resolution, background, \emph{etc}.

Our work is motivated by two aspects. First, local feature based approaches \cite{gray2008viewpoint, zhao2014learning, zhao2013person} are proven to be effective in person re-identification. Considering the ``query-search'' mode, this is potentially compatible with image search based on the Bag-of-Words (BoW) model. Nevertheless, some state-of-the-art methods in person re-identification rely on brute-force feature-feature matching \cite{zhao2013unsupervised, zhao2013person}. Although good recognition rate is achieved, this line of methods suffer from low computational efficiency, which limits its potential in large-scale applications.
%Our goal is to search in the gallery for images having the same ID as the query pedestrian.
In the BoW model, local features are quantized to \emph{visual words} using a pretrained \emph{codebook}. An image is thus represented by a visual word histogram weighted by TF-IDF scheme. Instead of performing exhaustive visual matching among images \cite{zhao2013unsupervised}, in the BoW model, local features are aggregated into a global vector. In tackling spatial constraints, a number of geometric-aware visual matching methods \cite{GVP, shen2012object, spatial_coding} are proposed. Moreover, to further boost search accuracy, it is beneficial to include some post-processing steps \cite{AKM, qin2011hello}.

\begin{figure*}
  \centering
  \includegraphics[width=6.8in]{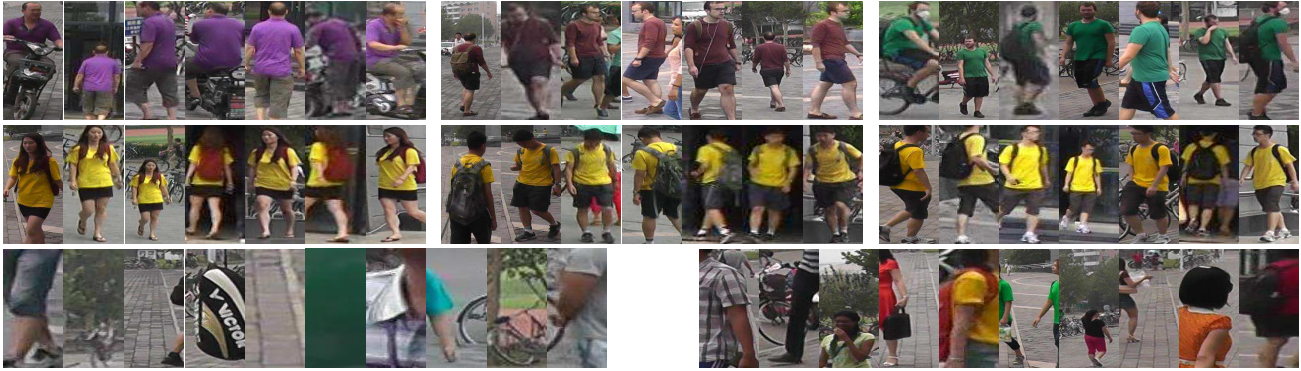}\\
  \caption{Sample images of the Market-1501 dataset. All images are normalized to 128$\times$64 (\textbf{Top:}) Sample images of three identities with distinctive appearance. (\textbf{Middle:}) We show three cases where three individuals have very similar appearance. (\textbf{Bottom:}) Some samples of the distractor images (left) as well as the junk images (right) are provided.}\label{fig:dataset_samples}
\end{figure*}
Second, most existing person re-identification datasets \cite{gray2007evaluating, Zheng2009associating, cheng2011custom, hirzer2012relaxed, li2013human, li2013locally} are flawed either in the dataset scale or in the data richness. Specifically, the number of identities is often confined in several hundred, which may lead to the performance instability. Moreover, images of the same identity are usually captured by two cameras; each identity typically has one image under each camera, so the number of queries and relevant images is very limited. Furthermore, in most datsets, pedestrians are well-aligned by hand-drawn bounding boxes. But in reality, when pedestrian detectors are employed, the detected persons may undergo misalignment or body part missing (see Fig. \ref{fig:dataset_samples}). On the other hand, pedestrian detectors, while producing true positive bounding boxes, also yield false alarms caused by complex background or occlusion (see also Fig. \ref{fig:dataset_samples}). In fact, these distractor images may exert non-ignorable influence on recognition accuracy. As a result, current methods may be biased toward more ideal settings and their effectiveness may be impaired once the ideal dataset meets reality. To address this problem, it is important to introduce datasets that reach closer to realistic settings and design robust algorithms which can handle detector errors and are not affected by distractors.

Considering the above two issues, this paper makes two major contributions. First, inspired by the state-of-the-art image search methods, an unsupervised BoW representation is proposed. After generating a codebook on the training data, each pedestrian image is represented as a visual word histogram. In this step, a number of techniques are integrated, \eg, root descriptor \cite{root_sift}, negative evidences \cite{jegou2012negative}, burstiness weighting \cite{burstiness}, \emph{etc}. To incorporate geometric constraints, images are partitioned into horizontal stripes. Moreover, multiple queries are pooled into one vector, which adapts to the extensive image variations. Finally, an automatic reranking step is added to refine the initial rank list. By simple dot product as similarity measurement, we show that the proposed feature representation yields competitive recognition accuracy while enjoying a fast response time.

Second, a new person re-identification dataset, called the ``Market-1501'', is introduced (Fig. \ref{fig:dataset_samples}). This dataset is composed of 1501 identities collected by 6 cameras near the entrance of a university campus supermarket. To the best of our knowledge, Market-1501 is the largest person re-identification dataset featured by 32643 annotated bounding boxes. It is distinguished from existing datasets in three aspects: DPM detected bounding boxes, the inclusion of distractor images, and multi-query, multi-groundtruth per identity. The Market-1501 dataset provides a more realistic benchmark for algorithm evaluation.

%a number of existing works \cite{prosser2010person, hirzer2011person, dikmen2011pedestrian, zheng2011person, li2014deepreid} leverage supervised approaches to learn discriminative features or metrics. This line of works are beneficial in reducing the impact of multi-view variations, but require laborious annotation, especially when new cameras are added in the system. This is intractable if we consider the large numbers of surveillance recorders city-wide. In this scenario, this paper proposes an unsupervised method which requires a minimal amount of labeling or training effort. Our system is well-adaptable to different camera networks, thus having higher generalization ability.

The rest of this paper is organized as follows. After a brief review of related works in Section \ref{section:related_work}, we describe the Market-1501 dataset in Section \ref{section:dataset_collection}. Then, Section \ref{section:method} introduces the proposed method based on image search techniques. Experimental results are summarized in Section \ref{section:experiments} and conclusions and insights are given in Section \ref{section:conclusion}.

\section{Related Work}\label{section:related_work}
For person re-identification, both supervised and unsupervised models have been extensively studied these years. In discriminative models \cite{prosser2010person, hirzer2011person, dikmen2011pedestrian, zheng2011person, li2014deepreid}, classic SVM (or the RankSVM \cite{prosser2010person, zhao2014learning}) and boosting \cite{gray2008viewpoint} are popular choices. For example, Zhao \etal \cite{zhao2014learning} learn the weights of filter responses and patch matching scores using RankSVM, and Gray \etal \cite{gray2008viewpoint} perform feature selection among an ensemble of local descriptors by boosting. Recently, li \etal \cite{li2014deepreid} propose a deep learning network to jointly optimize all pipeline steps. This line of works are beneficial in reducing the impact of multi-view variations, but require laborious annotation, especially when new cameras are added in the system. On the other hand, in unsupervised models, Farenzena \etal \cite{farenzena2010person} make use of both symmetry and asymmetry nature of pedestrians and propose the Symmetry-Driven Accumulation of Local Features (SDALF). Ma \etal \cite{ma2012local} employ the Fisher Vector to encode local features into a global vector. To exploit the salience information among pedestrian images, Zhao \etal \cite{zhao2013person} propose to assign higher weight to rare colors, an idea very similar to the Inverse Document Frequency (IDF) in image search. In this scenario, this paper proposes an unsupervised method which requires a minimal amount of labeling or training effort.

\setlength{\tabcolsep}{1.8pt}
\begin{table*}[t]
\centering
\begin{tabular}{|l|c|c|c|c|c|c|c|c|}
\hline
Datasets& Market-1501& RAiD \cite{das2014consistent} & CUHK03 \cite{li2014deepreid} & VIPeR \cite{gray2007evaluating} & i-LIDS \cite{Zheng2009associating} & CUHK01 \cite{li2013human}  & CUHK02 \cite{li2013locally} & CAVIAR \cite{cheng2011custom}\\

\hline
\hline
\# identities& 1,501& 43 & 1,360 & 632 & 119 & 971 & 1,816 & 72 \\
\hline
\# BBoxes& 32,643 &6920& 13,164 & 1,264 & 476 & 1,942 &  7,264 & 610\\
\hline
\# distractors& 2,793 &0& 0 & 0 & 0 & 0 &  0 & 0\\
\hline
\# cam. per ID & 6 & 4& 2 & 2 & 2 & 2 &  2 & 2 \\
\hline
DPM or Hand & DPM &hand& DPM & hand & hand & hand &  hand & hand\\
\hline
Evaluation & mAP &CMC& CMC & CMC & CMC & CMC &  CMC & CMC\\
\hline
\end{tabular}
\caption{Comparing Market-1501 with existing datasets \cite{li2014deepreid, gray2007evaluating, Zheng2009associating, li2013human, li2013locally, cheng2011custom}.}
\label{table:compare_datasets}
\end{table*}

On the other hand, the field of image search has been greatly advanced since the introduction of the SIFT descriptor \cite{SIFT2} and the BoW model. In the last decade, a myriad of methods \cite{Hamming, zheng2014packing, BOC, spatial_coding, GVP} have been developed to improve search performance. For example, to improve matching precision, J\'{e}gou \etal \cite{Hamming} embed binary SIFT features in the inverted file. Meanwhile, refined visual matching can also be produced by index-level feature fusion \cite{zheng2014packing, BOC} between complementary descriptors. Since the BoW model does not consider the spatial distribution of local features (also a problem in person re-identification), another direction is to model the spatial constraints \cite{spatial_coding, GVP, hoiem2004object}. The geometry-preserving visual phrases (GVP) \cite{GVP} and the spatial coding \cite{spatial_coding} methods both calculate the relative position among features, and check the geometric consistency between images by the offset maps. Zhang \etal \cite{DVP} propose to use descriptive visual phrases to build pairwise constraints, and Liu \etal \cite{liu2012embedding} encode geometric cues into binary features embedded in the inverted file.

%In this paper, to reduce the impact of misalignment, a multi-scale geometric matching method is introduced, which can be performed in an on-the-fly manner.

For ranking problems, an effective reranking step typically brings about improvements. Liu \etal \cite{liu2013pop} design a ``one shot'' feedback optimization scheme which allows a user to quickly refine the search results. Although it is shown to yield consistent improvement, reranking based on user feedback is not always desirable or accessible. In rigid object search, RANSAC \cite{AKM} is typically used in post-processing. In \cite{shen2012object}, the top-ranked images are used as queries again and final score is the weighted sum of individual scores. When multiple queries are present \cite{arandjelovic2012multiple}, a new query which integrates the original queries can be formed by average or max operations.

%Inspired by \cite{arandjelovic2012multiple}, in the Market-1501 dataset, since multiple ground truths exist, we add a reranking step by re-formulating the top-ranked images as a new query. We demonstrate consistent improvement on our dataset.

\section{The Market-1501 Dataset}\label{section:dataset_collection}
\subsection{Description}
In this paper, a new person re-identification dataset, the ``Market-1501'' dataset, is introduced. During dataset collection, a total of six cameras were placed in front of a campus supermarket, including five 1280$\times$1080 HD cameras, and one 720$\times$576 SD camera. Overlapping exists among these cameras. This dataset contains 32643 bounding boxes of 1501 identities. Since it is an open environment, images of each identity are captured by at most six cameras. We make sure that each annotated identity are captured by at least two cameras, so that cross-camera search can be performed. In fact, even within same camera, images of same identity still take on distinct appearance. Overall, the Market-1501 dataset has the following featured properties.

First, while most existing datasets use hand-cropped bounding boxes, the Market-1501 dataset employs a state-of-the-art detector, \ie, the Deformable Part Model (DPM) \cite{felzenszwalb2010object}.
%Based on the ``perfect'' hand-drawn bounding boxes, current methods do not fully consider the misalignment of pedestrian images, a problem which always exists in DPM based bounding boxes.
As is shown in Fig. \ref{fig:dataset_samples}, misalignment as well as body part missing are very common among the detected images.

Second, in addition to the false positive bounding boxes, we also provide false alarm detection results. We notice that the CUHK03 dataset \cite{li2014deepreid} also uses the DPM detector, but the bounding boxes in CUHK03 are relatively good ones in terms of DPM detector. In fact, a large number of the detected bounding boxes would be very ``bad''. Considering this, for each detected bounding box to be annotated, a hand-drawn groundtruth bounding box is provided (similar to \cite{li2014deepreid}). Different from \cite{li2014deepreid}, for the detected and hand-drawn bounding boxes, the ratio of the overlapping area to the union area is calculated. In our dataset, if the area ratio is larger than 50\%, the DPM bounding box is marked as ``good'' (a routine in object detection \cite{felzenszwalb2010object}); if the ratio is smaller than 20\%, the DPM bounding box is marked as ``distractor''; otherwise, the bounding box is marked as ``junk'' \cite{AKM}, meaning that this image is of zero influence to the re-identification accuracy. Moreover, some obvious false alarm bounding boxes on the background are also marked as ``distractors''. In Fig. \ref{fig:dataset_samples}, examples of ``good'' images are shown in the top two rows, while ``distractor'' and ``junk'' images are in the bottom row.

Third, each identity may have multiple images under each camera. Therefore, during cross-camera search, there may be multiple queries and multiple groundtruths for each identity. A comparison with existing datasets is shown in Table \ref{table:compare_datasets}.

\begin{figure}
  \centering
  \includegraphics[width=2.5in]{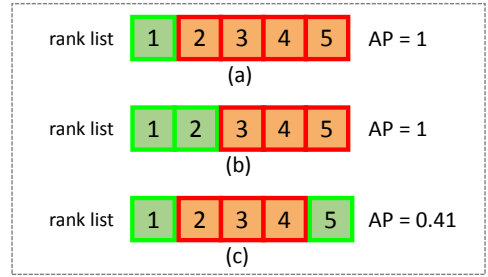}\\
  \caption{A toy example of the difference between AP and CMC measurements. True matches and false matches are in green and red boxes, respectively. For all three rank lists, the CMC curve remains 1. But AP = 1, 1, and 0.41, respectively.}\label{fig:map-cmc}
\end{figure}

\subsection{Evaluation Protocol}
Current datasets typically use the Cumulated Matching Characteristics (CMC) curve to evaluate the performance of re-identification algorithms. CMC curve shows the probability that a query identity appears in different sized candidate lists. This evaluation measurement is valid only if there is only one groundtruth match for a given query (see Fig. \ref{fig:map-cmc}(a)). In this case, the precision and recall are the same issue. However, if multiple groundtruths exist, the CMC curve is biased because ``recall'' is not considered. For example, CMC curves of Fig. \ref{fig:map-cmc}(b) and Fig. \ref{fig:map-cmc}(c) both equal to 1, which fails to provide a fair comparison of the quality between the two rank lists.

For Market-1501 dataset, there are on average 14.8 cross-camera groundtruths for each query. Therefore, we use mean average precision (mAP) to evaluate the overall performance. For each query, we calculate the area under the Precision-Recall curve, which is known as average precision (AP). Then, the mean value of APs of all queries, \ie, mAP, is calculated, which considers both precision and recall of an algorithm, thus providing a more comprehensive evaluation.

\begin{figure}
  \centering
  \includegraphics[width=2.8in]{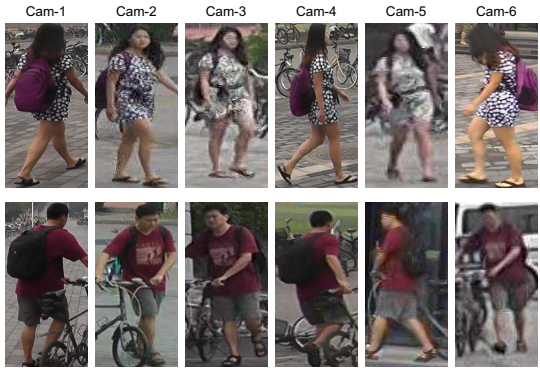}\\
  \caption{Samples query images. In Market-1501 dataset, queries are hand-drawn bounding boxes. Each identity has at most 6 queries, one for each camera.}\label{fig:query}
\end{figure}

Our dataset is randomly divided into training and testing sets, containing 750 and 751 identities, respectively. During testing, for each identity, we select one query image in each camera. Note that, the selected queries are hand-drawn, instead of DPM-detected as in the gallery. The reason is that in reality, it is very convenient to interactively draw a bounding box, which can yield higher recognition accuracy \cite{li2014deepreid}. The search process is performed in a cross-camera mode, \ie, relevant images captured in the same camera as the query are viewed as ``junk''. In this scenario, an identity has at most 6 queries, and there are 3363 query images in total. On average, there are 4.5 query images for each identity, and each query has 14.8 groundtruth images. Queries of two sample identities are shown in Fig. \ref{fig:query}.

\section{Our Method}\label{section:method}
\subsection{The Bag-of-Words Model}
For three reasons, we adopt the Bag-of-Words (BoW) model. First, this model well accommodates local features, which are indicated as effective by previous works \cite{ma2012local, zhao2013person}. Second, it enables fast global feature matching, instead of exhaustive feature-feature matching \cite{zhao2014learning, zhao2013unsupervised}. Third, by quantizing similar local descriptors to the same visual word, the BoW model achieves some invariance to illumination, view, \emph{etc}. We describe the individual steps below.

\begin{figure}
  \centering
  \includegraphics[width=2.7in]{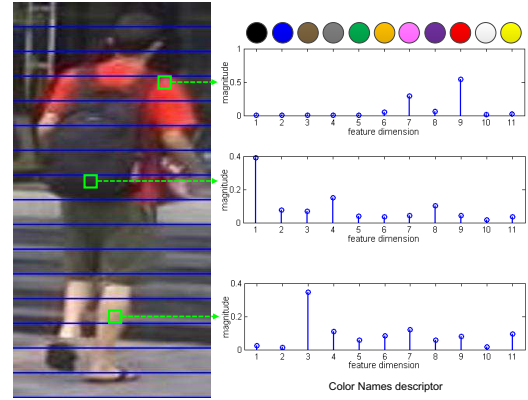}\\
  \caption{Local feature extraction. We compute the mean CN vector for each 4$\times$4 patch. These local features are quantized, and then pooled in a histogram for each horizontal stripe.}\label{fig:feature_extraction}
\end{figure}

\noindent\textbf{Feature Extraction.} As a baseline, we employ the Color Names (CN) descriptor \cite{van2009learning}. Given a pedestrian image normalized to 128$\times$64 pixels, patches of size 4$\times$4 are densely sampled. In our experiment, the sampling step is 4, so there is no overlapping between patches. For each patch, CN descriptors of all pixels are calculated, and are subsequently $\ell_1$ normalized followed by $\sqrt{(\cdot)}$ operator \cite{root_sift}. The mean vector is taken as the descriptor of this patch (see Fig. \ref{fig:feature_extraction}).

\noindent\textbf{Codebook.} For Market-1501 dataset, we generate a codebook on the training set. For other datasets, the codebook is trained on the independent TUD-Brussels dataset \cite{wojek2009multi}. Standard $k$-means is used, and codebook size is $k$.

\noindent\textbf{Quantization.} Given a local descriptor, we employ Multiple Assignment (MA) \cite{Hamming} to find its near neighbors under Euclidean distance in the codebook. We set MA = 10, so a feature is represented by the indices of 10 visual words.

\noindent\textbf{TF-IDF.} The visual word histogram is weighted by TF-IDF scheme. TF encodes the number of occurrences of a visual word, and IDF is calculated as $\log \frac{N}{n_i}$, where $N$ is the number of images in the gallery, and $n_i$ is the number of images containing visual word $i$. We use the avgIDF \cite{zheng2013lp} variant in place of the standard IDF.

\noindent\textbf{Burstiness.} Burstiness refers to the phenomenon where a query feature finds multiple matches in a test image \cite{burstiness}. For CN descriptor, burstiness could be more prevalent due to its low discriminative power compared with SIFT. Therefore, all terms in the histogram are divided by $\sqrt{tf}$.

\noindent\textbf{Negative Evidence.} Following \cite{jegou2012negative}, we calculate the mean feature vector in the training set. Then, the mean vector is subtracted from all test features. In this way, the zero entries in the feature vector are also taken into account using dot product.

\noindent\textbf{Similarity Function.} Given a query image $Q$ and a gallery image $G$, the corresponding $\ell_2$-normalized feature vectors are denoted as $(q_1, q_2,...,q_k)^T$ and $(g_1, g_2,...,g_k)^T$, respectively, where $k$ is codebook size. The similarity function is written as,
\begin{equation}\label{eq:sim_baseline}
  sim(Q, G) = \sum_{i = 1}^k q_i\cdot g_i,
\end{equation}
which calculates the dot product between two vectors.

\subsection{Improvements}
\noindent\textbf{Weak Geometric Constraints.} In image search, geometric clues among local features have been demonstrated as good discriminator to outlier matches \cite{shen2012object, Hamming, spatial_coding, AKM}. For person re-identification, popular approaches \cite{zhao2013person, zhao2014learning, zhao2013unsupervised} include the so-called ``Adjacency Constrained Search'' (ACS). In this method, a patch in the probe is matched with patches in a gallery image, which are located in a horizontal stripe positioned at similar height with the probe patch. The minimum distance is taken as the similarity score. Similar idea also appears in DeepReid \cite{li2014deepreid}.

ACS is effective in incorporating spatial constraints, but, as will be shown in the experiments, it suffers from high computational cost. Inspired by Spatial Pyramid Matching \cite{SPM}, we integrate ACS into the BoW model. As illustrated in Fig. \ref{fig:feature_extraction}, an input image is partitioned into $M$ horizontal stripes. Then, for stripe $m$, the visual word histogram is represented as $\bm{d}^{m} = (d_1^{m}, d_2^{m}, ..., d_k^{m})^T$, where $k$ is the codebook size. Consequently, the feature vector for the input image is denoted as $\bm{f} = (\bm{d}^{1}, \bm{d}^{2}, ...., \bm{d}^{M})^T$, which is the concatenation of vectors from all stripes. When matching two images, dot product (Eq. \ref{eq:sim_baseline}) is employed, which sums up the similarity at all corresponding stripes. In this manner, we avoid calculating patch distances for each query feature; instead, the calculation is performed at \emph{stripe level}.

\noindent\textbf{Background Suppression.} The negative impact of background distraction has been studied extensively \cite{farenzena2010person, zhao2013person, zhao2013unsupervised}. In one solution, Farenzena \etal \cite{farenzena2010person} propose to separate the foreground pedestrian from background using segmentation. In the following works, Zhao \etal \cite{zhao2013person, zhao2013unsupervised} use the resulting masks to filter out background.

Since the process of generating a mask for each image is both time-consuming and unstable, this paper proposes a simple solution by exerting a 2-D Gaussian template on the image. Specifically, the Gaussian function takes on the form of $N(\mu_x, \sigma_x, \mu_y, \sigma_y)$, where $\mu_x$, $\mu_y$ are horizontal and vertical Gaussian mean values, and $\sigma_x$, $\sigma_y$ are horizontal and vertical Gaussian standard variances, respectively. We set $(\mu_x, \mu_y)$ to the image center, and set $(\sigma_x, \sigma_y) = (1, 1)$ for all experiments. This method injects a prior knowledge on the position of pedestrian, which assumes that the person lies in the center, and is surrounded by background. Therefore, the Gaussian mask works by suppressing the response near the edge of the image.

\noindent\textbf{Multipe Queries.} The usage of multiple queries is shown to yield superior results in image search \cite{arandjelovic2012multiple} and re-identification \cite{farenzena2010person}. If we want to delimit a person, it would be worthy of using multiple query bounding boxes and re-formulating the query image. In this manner, the intra-class variation is taken into account, and the algorithm would be more robust to pedestrian variations.

When each identity has multiple query images in a single camera, instead of a multi-multi matching strategy \cite{farenzena2010person}, we merge them into a single query for speed consideration. Here, we employ two pooling strategies, \ie, average and max pooling. In average pooling, the feature vectors of multiple queries are pooled into one by averaged sum; in max pooling, the final feature vector takes the maximum value in each dimension from all queries.

\noindent\textbf{Automatic Reranking.}
When viewing person re-identification as a ranking problem, a natural idea consists in the usage of reranking algorithms. In this paper, we use a simple reranking method which picks top-$T$ ranked images of the initial rank list as queries to search the gallery again. Specifically, given an initial sorted rank list by query $Q$, image $R_i$ which is the $i^{th}$ image in the list is used as query. The similarity score of a gallery image $G$ when using $R_i$ as query is denoted as $S(R_i, G)$. We assign a weight $1/(i+1), i = 1,...,T$ to each top-$i$ ranked query, where $T$ is the number of expanded queries. Then, the final score of the gallery image $G$ to query $Q$ is determined as,
\begin{equation}\label{eq:rerank}
  \hat S(Q, G) = S(Q, G) + \sum_{i=1}^{T}\frac{1}{i+1}S(R_i, G),
\end{equation}
where $\hat S(Q, G)$ is the weighted sum of similarity scores obtained by the original and expanded queries, and the weight gets smaller as the expanded query is located away from the top. This method departs from the one proposed in \cite{shen2012object} in that Eq. \ref{eq:rerank} employs the similarity values while \cite{shen2012object} uses the reverse ranks.

\section{Experiments}\label{section:experiments}
\subsection{Datasets}
This paper experiments on three datasets, \ie, Market-1501, VIPeR \cite{gray2007evaluating}, and CUHK03 \cite{li2014deepreid}. The latter two datasets are described below.

\noindent\textbf{VIPeR.} This dataset is composed of 632 identities, and each has two images captured from two different cameras. Pedestrians in this dataset undergo large variances in viewpoint, illumination, pose, \emph{etc}. All images are normalized to 128$\times$48 pixels. VIPeR is randomly divided into two equal halves, one for training, and the other for testing. Each half contains 316 identities. For each identity, we take an image from one camera as query, and perform cross-camera search.

\noindent\textbf{CUHK03.} This dataset contains 13,164 DPM bounding boxes of 1,467 identities. Each identity is observed by two cameras and has 4.8 images in average for each view.
%CUHK03 dataset also demonstrates cases of misalignment, occlusion, and body part missing.
Following the protocol in \cite{li2014deepreid}, for the test set, we randomly select 100 persons. For each person, all the DPM bounding boxes are taken as query in turns, and a cross-camera search is performed. The test process is repeated 20 times, and stable statistics are reported. We report both the CMC scores and mAP for VIPeR and CUHK03 datasets.

\subsection{Important Parameters}
\noindent\textbf{Codebook size  $k$.} In our experiment, codebooks of various sizes are constructed, and mAP on Market-1501 dataset is presented in Table \ref{table:codebook_size}. We can see that the peak value is achieved when $k = 350$. In the following experiments, this value is kept.

\setlength{\tabcolsep}{10pt}
\begin{table}[!t]
\renewcommand{\arraystretch}{1}
\begin{center}
\begin{tabular}{|l|cccc|}
\hline
$k$& 100 & 200 & 350 &  500\\
\hline
mAP (\%)  & 13.31 &  14.01 & 14.09 &13.82\\
\hline
r=1 (\%)  & 32.20 & 34.24 & 34.40 &34.14\\
\hline
\end{tabular}
\caption{Impact of codebook size on Market-1501 dataset. We report results obtained by ``BoW + Geo + Gauss''.}
\label{table:codebook_size}
\end{center}
\end{table}

\setlength{\tabcolsep}{7pt}
\begin{table}[!t]
\renewcommand{\arraystretch}{1}
\begin{center}
\begin{tabular}{|l|ccccc|}
\hline
$M$ & 1 & 4 & 8 & 16 &32\\
\hline
mAP (\%)   & 5.23  & 11.01 & 13.26& 14.09& 13.79\\
\hline
r=1 (\%)   & 14.36 & 27.53 &32.50 &34.40& 34.58\\
\hline
\end{tabular}
\caption{Impact of number of horizontal stripes on Market-1501 dataset. We report results obtained by ``BoW + Geo + Gauss''.}
\label{table:stripes}
\end{center}
\end{table}

\setlength{\tabcolsep}{4.2pt}
\begin{table}[!t]
\renewcommand{\arraystretch}{1}
\begin{center}
\begin{tabular}{|l|cccccc|}
\hline
T & 0 & 1 & 2 &  3 &4&5\\
\hline
mAP (\%)  & 18.53 & 19.18  & 19.07 & 18.97& 19.01 & 18.91\\
\hline
\end{tabular}
\caption{Impact of number of expanded queries on Market-1501 dataset. $k=0$ corresponds to the ``BoW + Geo + Gauss + MultiQ\_max'' mode.}
\label{table:knn}
\end{center}
\end{table}

\noindent\textbf{Number of stripes $M$.} Table \ref{table:stripes} presents the performance of different numbers of stripes. As the stripe number increases, a finer partition of the pedestrian image leads to a more discriminative representation. So the recognition accuracy increases, but recall may drop for a large $M$. For example, $M = 32$ produces higher rank-1 matching rate but lower mAP than $M = 16$. As a trade-off between speed and accuracy, we choose to split an image into 16 stripes in our experiment.

\noindent\textbf{Number of expanded queries  $T$.}
%The performance of reranking relies on the selection of parameter $k$, \ie, the number of expanded queries in the initial rank list.
Table \ref{table:knn} summarizes the results obtained by different numbers of expanded queries. We find that the best performance is achieved when $T$ = 1. When $T$ increases, mAP drops slowly, which validates the robustness to $T$. The performance of reranking highly depends on the quality of the initial list, and a larger $T$ would introduce more noise. In the following, we set $T$ to 1.

\begin{figure}
  \centering
  \includegraphics[width=3.28in]{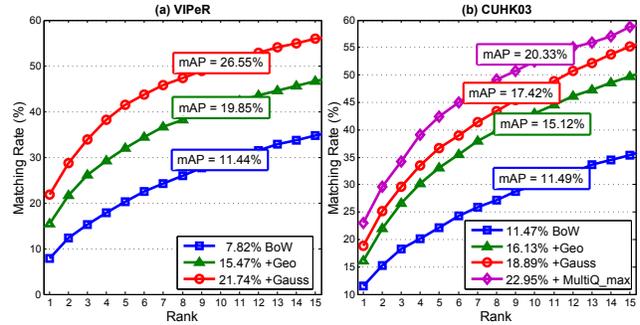}
  \caption{Performance of different method combinations on VIPeR and CUHK03 datasets.}\label{fig:methods_viper}
\end{figure}

\subsection{Evaluation}
\setlength{\tabcolsep}{9.5pt}
\begin{table*}[!t]
\renewcommand{\arraystretch}{1}
\begin{center}
\begin{tabular}{|l|cc|ccc|cc|}
\hline
\multirow{2}{*}{Methods} &
\multicolumn{2}{c|}{\emph{Market-1501}} &
\multicolumn{3}{c|}{\emph{VIPeR}} &
\multicolumn{2}{c|}{\emph{CUHK03}}\\
\cline{2-8}
& r = 1 & mAP & r = 1 &  r = 20 &mAP& r = 1 & mAP\\
\hline
\hline
BoW  & 9.04 & 3.26  & 7.82 &39.34&11.44&  11.47& 11.49\\
\hline
BoW + Geo & 21.23 &  8.46 & 15.47 &51.49&19.85& 16.13 & 15.12 \\
\hline
BoW + Geo + Gauss & 34.40 & 14.09 & 21.74 & 60.85 & 26.55& 18.89 &  17.42 \\
\hline
BoW + Geo + Gauss + MultiQ\_avg   & 41.66 & 17.63 & -& -& -& 22.35 & 20.48  \\
\hline
BoW + Geo + Gauss + MultiQ\_max   & 42.14 & 18.53 & -& -& -& 22.95 & 20.33 \\
\hline
BoW + Geo + Gauss + MultiQ\_max + Rerank & 42.14 & 19.20 & -&-& -& 22.95& 22.70\\
\hline
\end{tabular}
\caption{Results (rank-1, rank-20 matching rate, and mean Average Precision (mAP)) on three datasets by combining different methods, \ie, the BoW model (BoW), Weak Geometric Constraints (Geo), Background Suppression (Gauss), Multiple Queries by average (MultiQ\_avg) and max pooling (MultiQ\_max), and reranking (Rerank). Note that, here we use the Color Names descriptor for BoW.}
\label{table:various_approaches}
\end{center}
\end{table*}

\noindent\textbf{BoW model and its improvements.} On three datasets, we present results obtained by the BoW representation, geometric constraints, Gaussian mask, multiple queries, and reranking in Table \ref{table:various_approaches} and Fig. \ref{fig:methods_viper}. Five major conclusions can be drawn.

First, we find that the baseline BoW vector produces a relatively low accuracy: rank-1 matching rate = 9.04\%, 10.56\%, and 5.35\% on Market-1501, VIPeR, and CUHK03 datasets, respectively.

Second, when we integrate geometric constraints by stripe matching, we observe consistent improvement in recognition accuracy. On Market-1501 dataset, for example, mAP increases from 3.26\% to 8.46\% (+5.20\%), and an even larger improvement can be seen from rank-1 matching rate, from 9.04\% to 21.23\% (+12.19\%). On VIPeR dataset, the rank-1 matching rate rises from 9.95\% to 15.35\% (+5.40\%). The improvement on CUHK03 is very similar. Our results are consistent with previous studies \cite{zhao2013person, li2014deepreid} in that when narrowing the matching scope, matching precision can be improved.

Third, it is clear that the Gaussian mask works well on all three datasets. Specifically, we observe +5.63\% in mAP, +4.24\% in rank-1 matching rate, and +5.74\% in rank-1 matching rate on Market-1501, VIPeR, CUHK03 datasets, respectively. Therefore, the prior that pedestrian is located more or less in the center of the bounding box is statistically sound. Previous methods \cite{farenzena2010person} address this issue by isolating foreground from background, a method which may be influenced by complex background. Another possible direction consists in modeling the background in video. At this point, we plan to release the video together with the bounding boxes, so that a more precise foreground segmentation result can be produced.

\begin{figure}[t]
  \centering
  \includegraphics[width=3.0in]{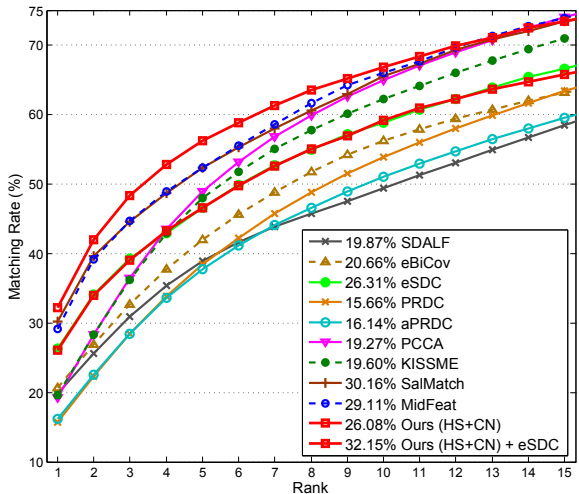}\\
  \caption{Comparison with the state-of-the-art methods on VIPeR dataset. For our method, we combine HS and CN features, as well as the eSDC method.}
  \label{fig:state-of-the-art}
\end{figure}
Then, we test multiple queries on CUHK03 and Market-1501 datasets, where each query identity has multiple bounding boxes. From the results, we can see that the usage of multiple queries further improves recognition accuracy. The improvement is more prominent on Market-1501 dataset, where the query images take on more diverse appearance (see Fig. \ref{fig:query}).
%On the contrary, queries of one identity in one camera are very similar on CUHK03 dataset, so their combination provides limited extra information.
Moreover, we notice that multi-query by max pooling is slightly superior to average pooling, probably because max pooling gives more weights to the rare but salient features and improves recall.

\setlength{\tabcolsep}{9pt}
\begin{table}[t]
\centering
\begin{tabular}{|l||ccc|}
\hline
Stage& SDALF & SDC & Ours \\

\hline
Feat. Extraction (s)& 2.92 & 0.76 & \bf0.62 \\
Search (s)& 2644.80 & 437.97 &\bf 0.98 \\
Rerank (s)& - & - &\bf 0.98 \\
\hline
\end{tabular}
\caption{Average query time of different steps on Market-1501 dataset. Our method achieves significant speedup.}
\label{table:query_time}
\end{table}

\setlength{\tabcolsep}{9.7pt}
\begin{table}[!t]
\renewcommand{\arraystretch}{1}
\begin{center}
\begin{tabular}{|l||c|cc|}
\hline
\multirow{2}{*}{Methods} &
\multicolumn{1}{c|}{CUHK03} &
\multicolumn{2}{c|}{Market-1501} \\
\cline{2-4}
& r = 1 & r = 1 & mAP \\
\hline
SDALF \cite{farenzena2010person} & 4.87 & 20.53 & 8.20 \\
ITML \cite{davis2007information} & 5.14 & - &  - \\
lMNN \cite{weinberger2005distance} & 6.25 & - &  - \\
eSDC \cite{zhao2013unsupervised} & 7.68 & 33.54 &  13.54 \\
RANK \cite{mcfee2010metric} & 8.52 &- &  - \\
LDM \cite{guillaumin2009you} & 10.92 & - & -  \\
KISSME \cite{kostinger2012large} & 11.70 & - & -  \\
FPNN \cite{li2014deepreid} & 19.89&-&-\\
\hline
Ours (no MultiQ) & 18.89 & 34.40 & 14.09  \\
Ours (MultiQ) & 22.95 & 42.14 & 19.20  \\
Ours (+HS) & \bf 24.33 & \bf 47.25 & \bf 21.88 \\
\hline
\end{tabular}
\caption{Comparison with the state-of-the-art methods on CUHK03 and Market-1501 datasets.}
\label{table:comp_state-of-the-art}
\end{center}
\end{table}

\begin{figure*}
  \centering
  \includegraphics[width=6.9in]{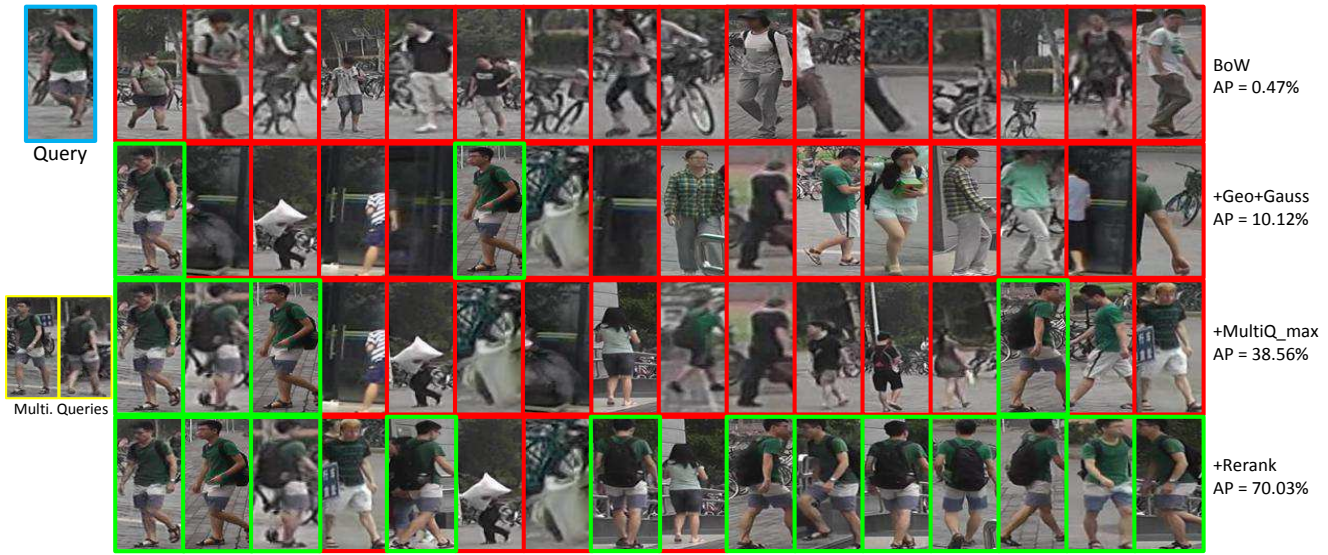}\\
  \caption{Sample re-identification results on Market-1501 dataset. Four rows correspond to four configurations, \ie, ``BoW'', ``BoW + Geo + Gauss'', ``BoW + Geo + Gauss + MultiQ'', and ``BoW + Geo + Gauss + MultiQ + Rerank''. The original query is in blue bounding box, and the added multiple queries are in yellow. Images with the same identity as the query is in green box, otherwise red. }\label{fig:sample_results}
\end{figure*}

Finally, we observe from Table \ref{table:knn} and Table \ref{table:various_approaches} that the reranking process generates higher mAP. Nevertheless, one recurrent problem with reranking is the sensitivity to the quality of initial rank list. On Market-1501 and CUHK03 datasets, since a majority of queries DO NOT have a top-1 match, the improvement in mAP is relatively small. For poor initial ranks, reranking would generate inferior results. Therefore, algorithms that produce higher accuracy may benefit more from reranking. Another tentative solution is to involve human interaction in the loop.\\

\noindent\textbf{Timings.} When the gallery gets scaled up (consider a city-scale re-identification system for an example), a fast response time is desirable. To evaluate this property, we compare our method with SDALF \cite{farenzena2010person} and SDC \cite{zhao2013unsupervised} in two aspects, \ie, feature extraction and search time.

Our evaluation is performed on a server with 3.46 GHz CPU and 128 GB memory, and efficiency results are shown in Table \ref{table:query_time}. We report the total timing by HS (we extract a 20-dim HS histogram and generate another BoW vector for fusion with CN) and CN features for our method. Compared with SDC, we achieve over two orders of magnitude efficiency improvement. In the SDALF method, three features are employed, \ie, MSCR, wHSV, and RHSP. The feature extraction time is 0.09s, 0.03s, 2.79s, respectively; the search time is 2643.94s, 0.66s, and 0.20s, respectively. Therefore, our method is faster than SDALF by three orders of magnitude. From these results, we can see that a major efficiency gain is achieved. \\

\noindent\textbf{Comparison with the state-of-the-arts.} We compare our results with the state-of-the-art methods in Fig. \ref{fig:state-of-the-art} and Table \ref{table:comp_state-of-the-art}. On VIPeR dataset (Fig. \ref{fig:state-of-the-art}), we first compare with unsupervised methods, \eg, eSDC \cite{zhao2013unsupervised}, SDALF \cite{farenzena2010person}. We can see that our approach is superior to both methods. Specifically, we achieve a rank-1 identification rate of 26.08\% when two features are used, \ie, Color Names (CN) and HS Histogram (HS). When eSDC \cite{zhao2013unsupervised} is further integrated, the matching rate increases to 32.15\%, a competitive accuracy among all competing methods.

Moreover, on CUHK03 dataset, our method without multiple-query (MultiQ) significantly outperforms almost all presented approaches. Compared with FPNN \cite{li2014deepreid} which builds a deep learning architecture, our accuracy is slightly lower by 1.00\%. But when multiple queries and HS feature are integrated, our rank-1 matching rate exceeds \cite{li2014deepreid} by +4.44\% on CUHK03 dataset. On Market-1501 dataset, compared with SDALF \cite{farenzena2010person}, eSDC \cite{zhao2013unsupervised}, and KISSME \cite{kostinger2012large}, our results are consistently higher.

Some sample results on Market-1501 dataset are provided in Fig. \ref{fig:sample_results}. Apart from the mAP increase with the method evolution, another finding which should be noticed is that the distractors detected by DPM on complex background or body parts severely affect re-identification accuracy. Previous works typically focus on ``good'' bounding boxes with person only, and rarely study the detector errors. In this sense, the Market-1501 dataset provides a more realistic environment for such evaluation.

\section{Conclusions and Insights}\label{section:conclusion}
This paper focuses on person re-identification. Overall, two major contributions are made. The first contribution consists in bridging the gap between person re-identification and BoW based image search. Specifically, the bag-of-words model with extensive improvements is applied, which considers the spatial constraints and the multi-query multi-groundtruth information. Second, a new person re-identification dataset, the Market-1501, is introduced. This dataset gets closer to the realistic settings, and once released, is one of the largest datasets in this field. Bounding boxes in the Market-1501 dataset are detected by DPM. Apart from annotated pedestrian images, we also include a number of false positive detection results, and view them as distractor or junk images.

The BoW representation, though unsupervised, achieves competitive results on three datasets, while speeding up the search process by over two orders of magnitude. Both are desirable properties in industrial usage and provide new perspectives in the field of person re-identification. We speculate that this model can be further improved in several directions. First, supervised approaches can be readily incorporated, \eg, RankSVM, metric learning, \emph{etc}, so that the global vector is more discriminatively weighted. This idea also works for multi-feature fusion, where descriptors such as VLAD \cite{indexing2}, CNN \cite{Jia13caffe}, can be effectively combined.
%Second, to further speeding up the search process, either the inverted file or hashing functions can be adopted.
Second, our experiment highlights the importance of geometric constraints and foreground estimation. In fact, the geometric cues can be more elaborately encoded \cite{GVP, spatial_coding}; the root and parts detected by DPM can be also incorporated. Moreover, the foreground can be more precisely located by modeling background statistics through video analysis. Finally, the strength of multiple queries can be further explored by SVM \cite{root_sift} or spatial verification \cite{AKM}. The Market-1501 dataset will be a useful benchmark enabling these research possibilities.

%The introduction of Market-1501 dataset will enable a variety of research topics in the future. For instance, we will explore how to fuse the detection confidence with the recognition scores. Also, supervised approaches can be employed for discriminative feature or metric learning. Specifically, the geometric matching kernel could be used for classifier training.

{\small
\bibliographystyle{ieee}
\bibliography{egbib}
}

\end{document}